\documentclass[10pt,twocolumn,letterpaper]{article}

\usepackage{wacv}

\makeatletter
\@namedef{ver@everyshi.sty}{}
\makeatother
\usepackage{tikz}

\usepackage{times}
\usepackage{epsfig}
\usepackage{graphicx}
\usepackage{amsmath}
\usepackage{amssymb}
\usepackage{booktabs}
\usepackage{algorithm}
\usepackage{algorithmic}
\usepackage{subcaption}
\usepackage{makecell}
\usepackage{multirow}

\def\wacvPaperID{924} 

\wacvalgorithmstrack   

\wacvfinalcopy 

\ifwacvfinal
\usepackage[breaklinks=true,bookmarks=false]{hyperref}
\else
\usepackage[pagebackref=true,breaklinks=true,colorlinks,bookmarks=false]{hyperref}
\fi

\begin{document}

\title{SVD-NAS: Coupling Low-Rank Approximation and Neural Architecture Search}

\author{Zhewen Yu,  Christos-Savvas Bouganis\\
Imperial College London\\
London, UK\\
{\tt\small \{zhewen.yu18, christos-savvas.bouganis\}@imperial.ac.uk}
}

\maketitle
\thispagestyle{empty}

\begin{abstract}
The task of compressing pre-trained Deep Neural Networks has attracted wide interest of the research community due to its great benefits in freeing practitioners from data access requirements. In this domain, low-rank approximation is a promising method, but existing solutions considered a restricted number of design choices and failed to efficiently explore the design space, which lead to severe accuracy degradation and limited compression ratio achieved. To address the above limitations, this work proposes the SVD-NAS framework that couples the domains of low-rank approximation and neural architecture search. SVD-NAS generalises and expands the design choices of previous works by introducing the Low-Rank architecture space, \textit{LR-space}, which is a more fine-grained design space of low-rank approximation. Afterwards, this work proposes a gradient-descent-based search for efficiently traversing the \textit{LR-space}. This finer and more thorough exploration of the possible design choices results in improved accuracy as well as reduction in parameters, FLOPS, and latency of a CNN model. Results demonstrate that the SVD-NAS achieves 2.06-12.85pp higher accuracy on ImageNet than state-of-the-art methods under the data-limited problem setting. SVD-NAS is open-sourced at \url{https://github.com/Yu-Zhewen/SVD-NAS}. 
\end{abstract}

\begin{figure*}
  \centering
  \includegraphics[width=\textwidth]{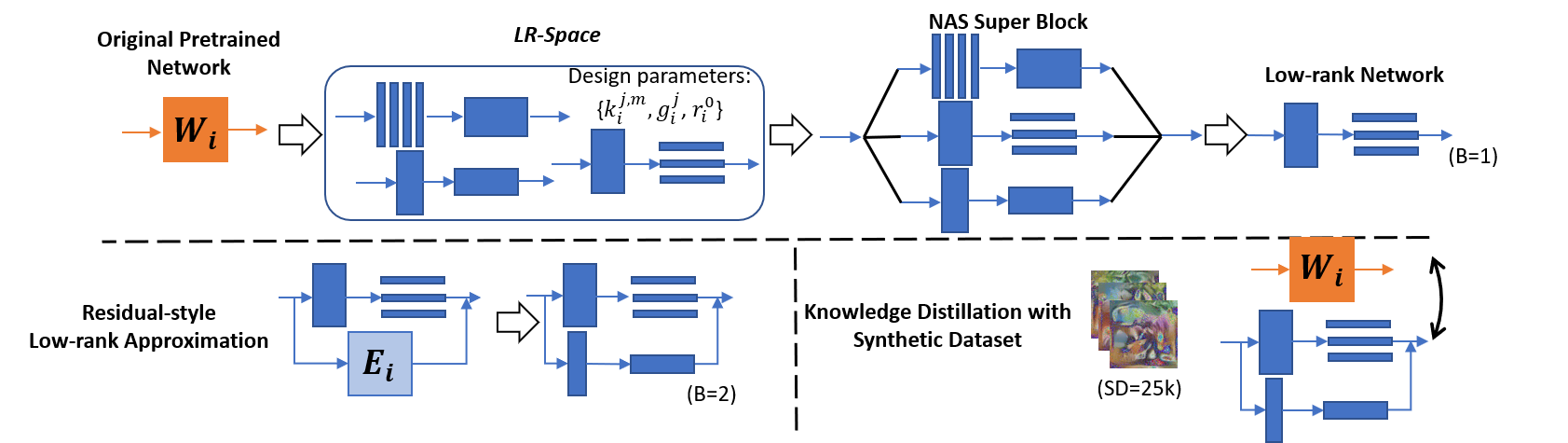}
  \caption{Main contributions of SVD-NAS. \textbf{Upper:} Given a pre-trained model, construct \textit{LR-space} and utilise NAS to identify optimal approximations. \textbf{Left Lower:} Extend \textit{LR-space} with the residual-style building block. \textbf{Right Lower:} Create a synthetic dataset and fine-tune the low-rank model by knowledge distillation.} \label{Fig:system_overview}
\end{figure*}

\section{Introduction}
\label{sec:Introduction}
Deep Neural Networks (DNNs) have attracted the interest of practitioners and researchers due to their impressive performance on a number of tasks, pushing the state-of-the-art beyond of what was thought to be achievable through classical Machine Learning methods. However, the high computational and memory storage cost of DNN models impede their deployment on resource-constrained edge devices. In order to produce lightweight models, the following techniques are often considered:
\begin{itemize}
\item compression of a pre-trained model followed by optional fine-tuning~\cite{lee2018snip,li2018constrained}.
\item compression-aware training, where the computational costs are integrated into the training objective as a regulariser ~\cite{idelbayev2020low}.
\item design and train a lightweight model by construction using domain knowledge or Automated Machine Learning (AutoML) \cite{sandler2018mobilenetv2,tan2019efficientnet}.
\end{itemize}

However, in the real-world scenario, the access to the original training dataset might not be easily granted, especially when the training dataset is of value or contains sensitive information. In this situation, compressing a pre-trained model has attracted wide interest of the research community, as the task of compression has the minimal requirement of data access.

Among the model compression methods, pruning \cite{lazarevich2021post} and quantisation \cite{banner2018post} have been well researched and deliver good results. However, the low-rank approximation approaches still remain a challenge on their application due to the severe accuracy degradation and limited compression ratio achieved \cite{liebenwein2021compressing}. The value of low-rank approximation originates from their potential impact on computational savings as well as their ability to result in structured computations, key element of today's computing devices.

This work considers the low-rank approximation problem of a Convolutional Neural Network (CNN). Let's consider a CNN that contains $L$ convolutional layers. Let's denote the weight tensor of the $i^{th}$ convolutional layer by  $\boldsymbol{W_{i}}$, where $i \in [0,L-1]$, and having dimensions $(f_{i}, c_{i}, k_{i}, k_{i})$, denoting $f_{i}$ filters, $c_{i}$ input channels and $k_{i} \times k_{i}$ kernel size. The low-rank approximation problem can be expressed as finding a set of low-rank tensors $\hat{\mathbb{W}}_{i} = \{\boldsymbol{\hat{W}_{i}^{0}}, \boldsymbol{\hat{W}_{i}^{1}}, ...\}$, and a function $F(\hat{\mathbb{W}}_{i})$ that approximate $\boldsymbol{W_{i}}$, in some metric space. Therefore, the low-rank approximation problem has two parts; to identify the decomposition scheme, i.e. the function $F$, and the rank kept to construct the low-rank tensors, i.e. $r_{i} = \{r_{i}^{0}, r_{i}^{1}, ...\}$, such that the metrics of interest are optimised. 

The above problem defines a large design space to be explored but existing approaches restrict themselves to only consider a small fraction of this space, by forcing the weight tensors in a network to adopt the same or similar decomposition schemes across all the layers in a network \cite{idelbayev2020low,li2019learning}. Moreover, even within their small sub-space, their design space exploration was slow and sub-optimum, either requires extensive hand-craft effort \cite{zhang2015accelerating} or is based on heuristics that employ the Mean Squared Error (MSE) of weight tensors approximations as a proxy of the network's overall accuracy degradation \cite{zhang2015efficient,yu2017compressing}.


In this work, we offer a new perspective in applying low-rank approximation, by converting it to a Neural Architecture Search (NAS) problem. The key novel aspects of this work are:

Firstly, we describe the process of low-rank approximation as a per-layer substitution of the original pre-trained network. For every layer, we introduce a Low-Rank architecture design space, \textit{LR-space}, which is defined by a set of parameterisable building blocks. We demonstrate searching the design parameters of these building blocks is equivalent to exploring different decomposition schemes and ranks. Afterwards, we utilise the gradient-descent NAS to navigate the \textit{LR-space}, and jointly optimise the accuracy and the computational requirements (e.g. FLOPs) of the compressed model. 

Secondly, a residual-style low-rank approximation is proposed to further refine the accuracy-FLOPs trade-off, based on a divide-and-conquer approach. We convert every convolutional layer in the original pre-trained network into the residual-style computational structure containing multiple branches, where each branch can have a distinct decomposition scheme and rank. Such a residual structure expands the design space, leading to a more fine-grained but still structured low-rank approximation solution.

Finally, motivated by previous work in model quantisation \cite{cai2020zeroq}, where the authors generated synthetic dataset to deal with the data-limited problem setting, we applied a similar approach for low-rank approximation. The synthetic dataset is fed into both the original pre-trained model and the compressed low-rank model, enabling the tuning of the weights of the compressed model through knowledge distillation, improving further the framework's performance without accessing the actual training data.

A comparison of the proposed framework to the state-of-the-art approach \cite{liebenwein2021compressing} demonstrates that our framework is able to achieve 2.06-12.85pp higher accuracy on ResNet-18, MobileNetV2 and EfficientNet-B0 while requiring similar or even lower FLOPs under the data-limited problem setting.

\section{Related Work}
\subsection{Low-rank Approximation}
Previous work on low-rank approximation of CNNs can be classified broadly into two categories depending on the underlying methodology applied; Singular Value Decomposition (SVD) and Higher-Order Decomposition (HOD). In the case of SVD, the authors of \cite{zhang2015accelerating,wang2017factorized,peng2018extreme} approximated $\boldsymbol{W_{i}}$ with two low-rank tensors where the latter tensor corresponds to a point-wise convolution. Their approaches differ in whether the first low-rank tensor corresponds to a grouped convolution and on the number of groups that it employs. Instead of having a point-wise convolution, Tai \textit{et al.} \cite{tai2015convolutional} implemented the low-rank tensors with two spatial-separable convolutions. Our framework uses the SVD algorithm for decomposing the weight matrix mainly because of its low complexity compared to HOD methods, such as Tucker \cite{kim2015compression} and CP \cite{lebedev2014speeding}, that use the more expensive Alternate Least Squares algorithm. 

\subsection{Neural Architecture Search}
NAS considers the neural network design process through the automatic search of the optimal combination of high-performance building blocks. The search can be performed using a top-down approach  \cite{cai2019once}, where a super network is initially trained and then pruned, or bottom-up \cite{liu2018darts} approach, where optimal building blocks are firstly identified and put together to form the larger network. Popular searching algorithms include reinforcement learning \cite{jiang2019accuracy}, evolutionary \cite{cai2019once} and gradient descent \cite{wu2019fbnet}. In this work, we adopt a gradient-descent search through a top-down approach to solve the low-rank approximation problem, and unlike the common problem setting of NAS which assumes the availability of large amount of training data, we focus on the data-limited scenario instead.

\begin{figure*}
  \centering
  \includegraphics[width=0.8\textwidth]{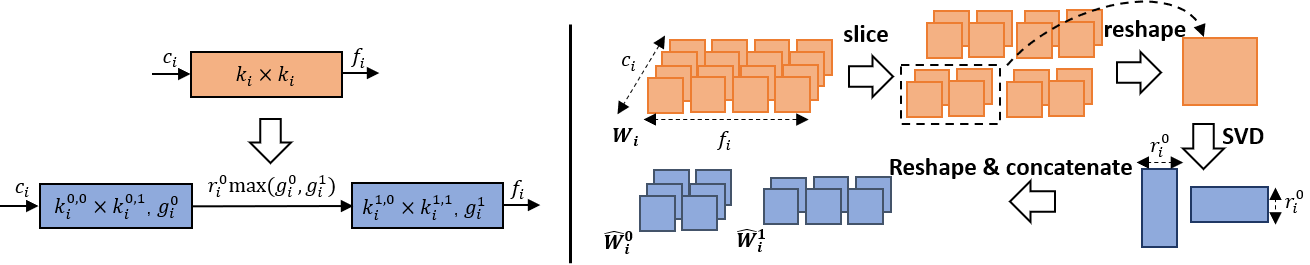}
    \caption{\textbf{Left:} Each building block contains two convolutional layers which are characterised by the design parameters: $k_{i}^{j,m}$, $g_{i}^{j}$ and $r_{i}^{0}$. \textbf{Right:} The process of deriving the low-rank weight tensors.} 
    \label{Fig:SVD_building_block}
\end{figure*}

\section{Design Space}
The objective of the proposed approach is to approximate each convolutional layer in a given CNN through a low-rank approximation such as the computation cost of the CNN is minimised, while a minimum penalty in the accuracy of the network is observed. Towards this, a design space, \textit{LR-space}, is firstly defined in this Section and a searching methodology to traverse that space will be introduced in Section \ref{sec:searching_algorithm}. 

\subsection{Low-rank Architecture Space (\textit{LR-space})}
\label{Subsec:SVD_Building_Blocks_and_LR-space}
In the rest of the paper, the convolutional layers in the pre-trained model are referred to as \textit{original layers}. Each \textit{original layer} is substituted with a parameterisable building block, as Fig.~\ref{Fig:SVD_building_block} Left shows. The building block has the same input and output feature map dimensions as the \textit{original layer}, but it contains two consecutive convolutional layers which are referred to as \textit{low-rank layers}.

The proposed building block is characterised through three design parameters: the low-rank kernel size $k_{i}^{j,m}$, the low-rank group number $g_{i}^{j}$ and the rank $r_{i}^{0}$, where $j \in \{0,1\}$, denoting the first and the second \textit{low-rank layer} respectively, and $m \in \{0,1\}$, denoting two spatial dimensions. In order to derive the weights of \textit{low-rank layers} from the \textit{original layer} with SVD-based decomposition, which will be elaborated in Subsection \ref{Subsec:SVD_Weight_Derivation}, additional constraints on the design parameters are introduced as follows:
\begin{itemize}
    \item $\prod\limits_{j}k_{i}^{j,m} = k_{i}$, which equivalently forces the kernel size of the \textit{low-rank layers} to be one of $\{1 \times 1$, $k_{i} \times k_{i}$, $k_{i} \times 1$, $1 \times k_{i}\}$, since $k_{i}$ is a prime number for most CNNs.
    \item $\underset{j}{min}(g_{i}^{j}) = 1$. This ensures that two \textit{low-rank layers} cannot be grouped convolutions at the same time.
    \item $r_{i}^{0}\underset{j}{max}(g_{i}^{j})(\frac{c_{i}}{g_{i}^{0}}\prod\limits_{m}k_{i}^{0,m}+\frac{f_{i}}{g_{i}^{1}}\prod\limits_{m}k_{i}^{1,m}) < c_{i}f_{i}k_{i}k_{i}$, the total number of weights inside two \textit{low-rank layers} should be less than the \textit{original layer}.
\end{itemize}

The proposed \textit{LR-space} generalises previous works which only took a subset of the space into account. Specifically, \cite{zhang2015accelerating,wang2017factorized,denton2014exploiting} considered the corner case of low-rank group number that $g_{i}^{0} \in \{1,f_{i},c_{i}\}$. Although \cite{peng2018extreme,li2019learning} introduced a design parameter to control group number, they did not explore different possibilities of kernel sizes as well as not attempt to put the grouped convolution in the second layer.

\subsection{Data-free Weight Derivation}
\label{Subsec:SVD_Weight_Derivation}
In this Subsection, we demonstrate how to use SVD to derive the weights of \textit{low-rank layers} from the \textit{original layers} in a data-free manner. Same as before, we denote the weight tensor of the \textit{original layer} by $\boldsymbol{W_{i}}$, while the weight tensors of the corresponding two \textit{low-rank layers} are denoted by $\boldsymbol{\hat{W}_{i}^{0}}$ and $\boldsymbol{\hat{W}_{i}^{1}}$ respectively.

$\boldsymbol{W_{i}}$, as a 4-d tensor, has the dimensions of $(f_{i}, c_{i}, k_{i}, k_{i})$. As \eqref{equ:svd_slice} shows, if we slice and split $\boldsymbol{W_{i}}$ on its first and second dimension into $g_{i}^{1}$ and $g_{i}^{0}$ groups respectively, we will obtain $g_{i}^{0}g_{i}^{1}$ tensors in total, where the dimensions of each tensor are $(\frac{f_{i}}{g_{i}^{1}}, \frac{c_{i}}{g_{i}^{0}}, k_{i}, k_{i})$.

\begin{multline}
    \boldsymbol{W_{i, q, p}} = \boldsymbol{W_{i}}[q\frac{f_{i}}{g_{i}^{1}}:(q+1)\frac{f_{i}}{g_{i}^{1}},\;p\frac{c_{i}}{g_{i}^{0}}:(p+1)\frac{c_{i}}{g_{i}^{0}},\;:,\;:], \\  p \in [0,g_{i}^{0}-1],\;q \in [0,g_{i}^{1}-1]     \label{equ:svd_slice}
\end{multline}

Due to the previous constraint on design parameters, we can substitute $k_{i}$ with the low-rank kernel size $k_{i}^{j,m}$. Therefore, the dimensions of each sliced tensor can also be expressed as $(\frac{f_{i}}{g_{i}^{1}}, \frac{c_{i}}{g_{i}^{0}}, k_{i}^{0,0}k_{i}^{1,0}, k_{i}^{0,1}k_{i}^{1,1})$. If we now reshape each sliced tensor $\boldsymbol{W_{i, q, p}}$ from 4-d to 2-d, we obtain the tensors $\mathcal{W}_{i, q, p}$, each having the dimensions of $(\frac{f_{i}}{g_{i}^{1}}k_{i}^{1,0}k_{i}^{1,1}, \frac{c_{i}}{g_{i}^{0}}k_{i}^{0,0}k_{i}^{0,1})$.

Applying SVD to $\mathcal{W}_{i, q, p}$ and keeping only the top-$r_{i}^{0}$ singular values, we obtain the following approximation,

\begin{equation}
    \mathcal{W}_{i, q, p} = USV \approx U_{r_{i}^{0}} S_{r_{i}^{0}} V_{r_{i}^{0}} = \mathcal{\hat{W}}_{i, q, p}^{1}\mathcal{\hat{W}}_{i, q, p}^{0}
\end{equation}
where $\mathcal{\hat{W}}_{i, q, p}^{0}$ and $\mathcal{\hat{W}}_{i, q, p}^{1}$ are 2-d low-rank tensors after absorbing the truncated diagonal matrix $S_{r_{i}^{0}}$ into $V_{r_{i}^{0}}$ and $U_{r_{i}^{0}}$.

The obtained 2-d low-rank tensors are reshaped back into the 4-d weight tensors, and they are concatenated together on their first and second dimension, which reverts the slice operation in \eqref{equ:svd_slice}. Eventually, two 4-d low-rank weight tensors are generated, denoted by $\boldsymbol{\hat{W}_{i}^{0}}$ and $\boldsymbol{\hat{W}_{i}^{1}}$, and have the dimensions of $(r_{i}^{0}, \frac{c_{i}}{g_{i}^{0}}, k_{i}^{0,0}, k_{i}^{0,1})$ and $(\frac{f_{i}}{g_{i}^{1}}, r_{i}^{0}, k_{i}^{1,0}, k_{i}^{1,1})$ respectively.

Recall that the SVD-based low-rank approximation problem is to identify optimal $F(\boldsymbol{\hat{W}_{i}^{1}}, \boldsymbol{\hat{W}_{i}^{0}})$ that approximates $\boldsymbol{W_{i}}$, involving choosing both the decomposition scheme and the decomposition rank. Among the design parameters of \textit{LR-space}, $k_{i}^{j,m}$ and the $g_{i}^{j}$ determine how the slicing and reshaping are performed, which correspond to the decomposition scheme $F$, while $r_{i}^{0}$ represents the decomposition rank.

\subsection{Residual Extension of \textit{LR-space}}
We also propose a residual-style building block as an extension of the \textit{LR-space} in order to further refine metrics trade-off. Continuing the previous analysis on the weight tensors, the process of low-rank approximation  replaces $\boldsymbol{W_{i}}$ with $\boldsymbol{\hat{W}_{i}^{1}}$, $\boldsymbol{\hat{W}_{i}^{0}}$, and injects the error $\boldsymbol{E_{i}}$ at the same time.

\begin{equation}
    \boldsymbol{W_{i}} = F(\boldsymbol{\hat{W}_{i}^{1}}, \boldsymbol{\hat{W}_{i}^{0}}) + \boldsymbol{E_{i}}
    \label{Equ:residual}
\end{equation}

So far, $\boldsymbol{E_{i}}$ is completely ignored and pruned. Alternatively, we can choose to keep part of $\boldsymbol{E_{i}}$ by further applying low-rank approximation to $\boldsymbol{E_{i}}$. Therefore, 

\begin{equation}
    \boldsymbol{W_{i}} = \sum_{b=0}^{1}F_{b}(\boldsymbol{\hat{W}_{i}^{1,b}}, \boldsymbol{\hat{W}_{i}^{0,b}})
\end{equation}

which corresponds to a residual-style building block with 2 branches whose outputs are element-wise summed, each branch containing two unique \textit{low-rank layers}. The superscript $b$ is to distinguish these two branches. The computation in the first branch is to approximate $\boldsymbol{W_{i}}$ while the second branch is to approximate $\boldsymbol{E_{i}}$. Although both branches have been low-rank approximated, their decomposition schemes and decomposition ranks can differ with each other, which makes the low-rank approximation more fine-grained compared with merely having one branch and increasing its rank.

\section{Searching Algorithm}
\label{sec:searching_algorithm}
Having defined the design space, the proposed framework considers the following multi-objective optimisation problem. The formulation aims to minimise the required number of computations per layer and at the same time to maximise the achieved accuracy of the network, subject to the form of decomposition.
\begin{multline}
    \min_{k_{i,b}^{j,m}, g_{i,b}^{j}, r_{i,b}^{0}} FLOP(\boldsymbol{\hat{W}_{i,b}^{j}}),\; \max_{k_{i,b}^{j,m}, g_{i,b}^{j}, r_{i,b}^{0}} ACC(\boldsymbol{\hat{W}_{i,b}^{j}}), \\ i \in [0,L-1], j \in \{0,1\}, m \in \{0,1\}, b \in \{0,1\} \label{equ:nas_target}
\end{multline}

$FLOP$ and $ACC$ represents the total operations and validation accuracy of the low-rank model respectively. 

\subsection{Gradient-descent NAS}
The framework uses a standard gradient-descent NAS approach \cite{wu2019fbnet} to solve the above optimisation problem. As Fig.~\ref{Fig:system_overview} shows, each convolutional layer in the pre-trained model is replaced with a super block during the search. The per-layer super block is constructed by exhaustively traversing through all the combinations of the design parameters of the \textit{LR-space} and instantiating the corresponding building blocks. Notice that the \textit{original layer} is also included as a candidate inside the super block, which provides the option to not compress this layer at all. 

The super block provides a Gumbel-Softmax weighted sum of the candidate building blocks drawn from \textit{LR-space}. Within this weighted sum, the weight of each candidate is given by $\boldsymbol{\theta_{i}}$, which is known as the sampling parameter in the previous literature to be distinguished from $\hat{\mathbb{W}}_{i}$, the actual weight tensors of convolution. During the search, the sampling parameter $\boldsymbol{\theta_{i}}$ gets updated with gradient descent by minimising the following multi-objective loss function.
\begin{equation}
    l_{nas}(\boldsymbol{\theta}) = l_{ce} \cdot [log(FLOP_{\hat{\mathbb{N}}})/log(FLOP_{\mathbb{N}})]^{\beta}
    \label{equ:nas_loss}
\end{equation}

where $l_{ce}$ is the cross-entropy loss, while $FLOP_{\hat{\mathbb{N}}}$
and $FLOP_{\mathbb{N}}$ represents the FLOPs of the compressed model and the original model respectively. $\beta$ is a hyperparameter which implicitly controls the compression ratio. When calculating the FLOPs of the super block, we also take the weighted sum of each candidate by the sampling parameter. At the end of the search, the candidate with the largest value of the sampling parameter is finally selected to replace the \textit{original layer}.

\subsection{Reduce Searching Cost}
\label{subsec:Reduce_Searching_Cost}
It is well-known that NAS can be very time-consuming and GPU-demanding given the huge size of the design space to be explored. For example, considering the \textit{LR-space} of a single convolutional layer where $(f_{i}, c_{i}, k_{i}, k_{i})$ is $(64, 64, 3, 3)$, there are 74902 candidates to be compared for approximating that layer. In this Subsection, we introduce some techniques to help explore the design space more efficiently, but at the same time, we can still keep the design choices of our framework more fine-grained than the previous work. 

Before the searching starts, we prune the \textit{LR-space} to reduce the number of candidate configurations considered by the framework. The following strategies are considered:

\begin{itemize}
    \item prune by FLOPs, we perform a grid search across the FLOPs. For example, we are only interested in those candidates whose FLOPs is close to $\{95\%, 90\%, 85\%, ...\}$ of the \textit{original layer}.  
    \item prune by accuracy, we use a proxy task where only one layer from the original network is compressed using the candidate configuration, while all other layers are left uncompressed, and we evaluate the corresponding accuracy degradation. The candidate will be pruned from the design space if this degradation is larger than a pre-defined threshold $\tau_{proxy}$. 
\end{itemize}

During the searching, we explore the space by an iterative searching method, which only searches for the configuration of one branch each time rather than simultaneously. We start with the case that there is only one branch and no residual structure inside the building block. With the help of NAS, we find the optimal configuration of design parameters belonging to that branch and fix that configuration. Afterwards, we add the residual branch into the building block and we start the searching again. 

Moreover, during every forward pass of searching, we sample and only compute the weighted sum of two candidates rather than all of them. The probabilities of each candidate getting sampled are the softmaxed $\boldsymbol{\theta_{i}}$. This technique was proposed by \cite{cai2018proxylessnas} to reduce GPU memory.

\begin{algorithm}[h]
\caption{Iterative Searching}
\label{alg:low_rank_approximation_of_residual_svd}
\small
\begin{algorithmic}[1]
\STATE $\boldsymbol{E_{i,0}} = \boldsymbol{W_{i}}$
\FOR{$b \in \{0, 1\}$}
\STATE identify optimal $F_{b}(\boldsymbol{\hat{W}_{i}^{1,b}}, \boldsymbol{\hat{W}_{i}^{0,b}})$ to approximate $\boldsymbol{E_{i,b}}$
\STATE $\boldsymbol{E_{i,b+1}} = \boldsymbol{E_{i,b}}-F_{b}(\boldsymbol{\hat{W}_{i}^{1,b}}, \boldsymbol{\hat{W}_{i}^{0,b}})$
\ENDFOR
\end{algorithmic}
\end{algorithm}

\section{Experiments}
The proposed SVD-NAS framework is evaluated on the ImageNet dataset using pre-trained ResNet-18, MobileNetV2 and EfficientNet-B0 coming from torchvision\footnote{https://github.com/pytorch/vision}. Thanks to techniques discussed in \ref{subsec:Reduce_Searching_Cost}, we can perform the searching on a single NVIDIA GeForce RTX 2080 Ti or a GTX 1080 Ti. Details of hyperparameters set-up can be found in this paper's supplementary material. 

\subsection{Performance Comparison}
According to the previous work \cite{banner2018post,migacz20178, li2020few}, the data-limited problem setting can be interpreted as two kinds of experiment set-up: post-training, where no training data is allowed for fine-tuning, and few-sample training, only a tiny subset of training data can be used for fine-tuning. 

For both set-ups, the proposed SVD-NAS framework's performance was evaluated and compared against existing works on CNN compression. The metrics of interest include, the reduction of FLOPs and parameters, in percentage (\%), as well as the degradation of Top-1 and Top-5 accuracy, in percentage point (pp).

\subsubsection{Post-training without tuning} 
We firstly report the performance of the compressed model without any fine-tuning. Table~\ref{tab:post_training_low_rank_approximation} presents the obtained results of the proposed framework for a number of networks and contrasts them to current state-of-the-art approaches. 

ALDS \cite{liebenwein2021compressing} and LR-S2 \cite{idelbayev2020low} are two automatic algorithms based on the MSE heuristic, while F-Group \cite{peng2018extreme} is a hand-crafted design. The results show that SVD-NAS outperforms all existing works when no fine-tuning is applied. In more details, on ResNet-18 and EfficientNet-B0, our work produced designs that achieve the highest compression ratio in terms of FLOPs as well as parameters, while maintaining a higher accuracy than other works. In terms of MobileNetV2, we achieve the best accuracy-FLOPs trade-off but not the best parameters reduction, as we do not include the number of parameters as an objective in \eqref{equ:nas_loss}.

\subsubsection{Post-training, but tuning with synthetic data}
Even though our framework outperforms the state-of-the-art approaches, we still observe a significant amount of accuracy degradation when no fine-tuning is applied. As training data are not available in the post-training experiment set-up, the proposed framework considers the generation of an unlabelled synthetic dataset and then uses knowledge distillation to guide the tuning of the parameters of the obtained model. 

Inspired by the previous work on post-training quantisation \cite{cai2020zeroq}, the synthetic data are generated by optimising the following loss function on the randomly initialised image $\boldsymbol{I}$:
\begin{multline}
    l_{bn}(\boldsymbol{I}) = \alpha[(\mu_{\boldsymbol{I}}')^2+(\sigma_{\boldsymbol{I}}'-1)^2] \\+ \sum_{i=0}^{L-1}\frac{1}{f_{i}}\sum_{f=0}^{f_{i}-1}[(\mu_{f}'-\mu_{f})^2+(\sigma_{f}'-\sigma_{f})^2]
     \label{Eq:bn_loss}
\end{multline}

$\mu_{f}$ and $\sigma_{f}$ is the running mean and running standard deviation stored in the batch normalisation layers from the pre-trained model. $\mu_{f}'$ and $\sigma_{f}'$ represents the corresponding statistics recorded when the current image is fed into the original pre-trained network. In addition, $\mu_{\boldsymbol{I}}'$ and $\sigma_{\boldsymbol{I}}'$ represents the mean and standard deviation of the current image itself, while $\alpha$ is a hyperparameter that balances these two terms.

Once the synthetic dataset is generated, we treat the original pre-trained model as the teacher and the compressed low-rank model as the student. Since the synthetic dataset is unlabelled, the knowledge distillation focuses on minimising the MSE of per-layer outputs. According to the results in Table~\ref{tab:post_training_low_rank_approximation}, the synthetic dataset can improve the top-1 accuracy by 2.44pp-7.50pp on three targeted models, which enlarges our advantage over the state-of-the-art methods.

\subsubsection{Few-Sample Training}
Few-sample training differs from the previous post-training in that now a small proportion of training data are available for the fine-tuning purpose. Specifically, for evaluation, we randomly select 1k images from the ImageNet training set as a subset and fix it throughout the experiment. During the fine-tuning, we use the following knowledge distillation method,

\begin{multline}
    l_{kd}(\hat{\mathbb{W}}_{i}) = \sum_{i=0}^{L-1}MSE(\hat{y_{i}}, y_{i}) \\+ \alpha_{kd} \cdot T_{kd}^2 \cdot l_{KLdiv} + (1-\alpha_{kd}) \cdot l_{ce}  
    \label{equ:kd_few_sample}
\end{multline}

where $MSE(\hat{y_{i}}, y_{i})$ stands for the Mean Square Error on the outputs of convolutional layers, while $l_{KLdiv}$ is the KL divergence on logits which are softened by temperature $T_{kd}$ (set as 6). $l_{ce}$ is the cross-entropy loss on the compressed model. Hyperparameter $\alpha_{kd}$ is set as 0.95. 

As none previous work has reported any result on few-sample low-rank approximation, we compare our framework with existing works on few-sample pruning instead. From Table \ref{tab:comparison_with_few_sample_pruning.}, our SVD-NAS framework provides a competitive accuracy-FLOPs trade-off on ResNet-18, especially when we are interested in those structured compression methods. We also observe that the compression ratio of MobileNetV2 achieved by our method is relatively less profound than the pruning methods, as that network contains a lot of depth-wise and point-wise convolutions which are less redundant in terms of the ranks of weight tensors.

\begin{table}
\footnotesize
\setlength\tabcolsep{1pt}
\begin{center}
\caption{Post-training results of low-rank approximation. $^{*}$ no fine-tuning. $^{**}$ fine-tuning with 25k synthetic images}
\label{tab:post_training_low_rank_approximation}
\begin{tabular}{llllll}
\hline\noalign{\smallskip}
Model &  Method & \makecell{$\Delta$ FLOPs \\ (\%)} & \makecell{$\Delta$ Params\\ (\%)}  & \makecell{$\Delta$ Top-1\\ (pp)} & \makecell{$\Delta$ Top-5\\ (pp)}\\
\noalign{\smallskip}
\hline
\noalign{\smallskip}
\multirow{5}*{ResNet-18} & \multirow{2}*{\bf SVD-NAS} & \multirow{2}*{\bf -58.60} & \multirow{2}*{\bf -68.05} & $-13.35^{*}$ &  $-9.14^{*}$ \\
~ & ~ & ~ & ~ & $\bf -5.85^{**}$ & $\bf -3.34^{**}$\\
\cline{2-6}
~ & ALDS \cite{liebenwein2021compressing} & -42.31 & -65.14 & -18.70 & -13.38\\
~ & LR-S2 \cite{idelbayev2020low} & -56.49 & -57.91 & -38.13 & -33.93\\
~ & F-Group\cite{peng2018extreme} & -42.31 & -10.66 & -69.34 & -87.63\\
\hline
\multirow{4}*{MobileNetV2} & \multirow{2}*{\bf SVD-NAS} & \multirow{2}*{\bf -12.54} & \multirow{2}*{-9.00} & $-15.09^{*}$ & $-7.79^{*}$ \\
~ & ~ & ~ & ~ & $\bf -9.99^{**}$ & $\bf -6.11^{**}$\\
\cline{2-6}
~ & ALDS \cite{liebenwein2021compressing} & -2.62 & \bf -37.61 & -16.95 & -10.91 \\
~ & LR-S2 \cite{idelbayev2020low} & -3.81 & -6.24 & -17.46 & -10.34 \\
\hline
\multirow{4}*{EfficientNet-B0} & \multirow{2}*{\bf SVD-NAS} & \multirow{2}*{\bf -22.17} & \multirow{2}*{\bf -16.41} & $-10.11^{*}$ & $-5.49^{*}$\\
~ & ~ & ~ & ~ & $\bf -7.67^{**}$ & $\bf -4.06^{**}$\\
\cline{2-6}
~ & ALDS \cite{liebenwein2021compressing} & -7.65 & -10.02 & -16.88 & -9.96 \\
~ & LR-S2 \cite{idelbayev2020low} & -18.73 & -14.56 & -22.08 & -14.15\\
\hline
\end{tabular}
\end{center}
\end{table}

\begin{table}
\footnotesize
\setlength\tabcolsep{1pt}
\begin{center}
\caption{Comparison with few-sample pruning.}
\label{tab:comparison_with_few_sample_pruning.}
\begin{tabular}{lllllll}
\hline\noalign{\smallskip}
Model &  Method & Struct. & \makecell{$\Delta$ FLOPs \\ (\%)} & \makecell{$\Delta$ Params\\ (\%)}  & \makecell{$\Delta$ Top-1\\ (pp)} & \makecell{$\Delta$ Top-5\\ (pp)}\\
\noalign{\smallskip}
\hline
\noalign{\smallskip}
\multirow{4}*{ResNet-18} &\bf SVD-NAS& yes& \bf -59.17 & \bf -66.77 & \bf -3.95 & -2.36 \\
\cline{2-7}
~ & FSKD \cite{li2020few} & yes & -59.01 & -64.64 & -6.01 & -\\
~ & Reborn \cite{tang2020reborn} & yes & -33.33 & - & - & -4.24\\
~ & POT \cite{lazarevich2021post} & no & - & -50.00 & - & \bf -1.48\\ 
\hline
\multirow{3}*{MobileNetV2} & \bf SVD-NAS& yes & \bf -14.17 & -10.66 & -6.63  &-3.61 \\
\cline{2-7}
~ & MiR \cite{wang2022compressing} & yes & -13.30 &  -7.70 & \bf -1.80 & - \\
~ & POT \cite{lazarevich2021post} & no & - & \bf -40.00 & - & \bf -2.87\\ 
\hline
\end{tabular}
\end{center}
\end{table}

\subsubsection{Full Training}
Although we are mainly interested in the data-limited scenarios, it is also interesting to remove the constraint of data availability and check the results when the full training set is available. Under this setting, 
we totally abandon knowledge distillation and only keep the cross-entropy term in \eqref{equ:kd_few_sample} for fine-tuning. All other experiment settings remain the same as before. 

Table~\ref{tab:fine_tune_low_rank_network_on_the_whole_training_set} presents the obtained results. In the case of ResNet-18, SVD-NAS reduces 59.17\% of the FLOPs and 66.77\% of parameters without any penalty on accuracy. In the case of MobileNetv2, the proposed framework produces competitive results as the other state-of-the-art works.

To summarise, we observe that the advantage of our framework over SOTA is correlated with the problem setting on data availability, as the advantage is more prominent in post-training and few-sample training, but is less evident in full training. This finding suggests that when the data access is limited, the design choices of low-rank approximation should be more carefully considered, while when a lot of training data are available, the performance gap between different design choices can be compensated through fine-tuning. 

\begin{table}
\footnotesize
\setlength\tabcolsep{1pt}
\begin{center}
\caption{Fine-tune low-rank network on the full training set.}
\label{tab:fine_tune_low_rank_network_on_the_whole_training_set}
\begin{tabular}{llllll}
\hline\noalign{\smallskip}
Model &  Method & \makecell{$\Delta$ FLOPs \\ (\%)} & \makecell{$\Delta$ Params\\ (\%)}  & \makecell{$\Delta$ Top-1\\ (pp)} & \makecell{$\Delta$ Top-5\\ (pp)}\\
\noalign{\smallskip}
\hline
\noalign{\smallskip}
\multirow{7}*{ResNet-18} & \bf SVD-NAS & -59.17 & -66.77& \bf +0.03 & \bf +0.10 \\
\cline{2-6}
~ & ALDS \cite{liebenwein2021compressing} & -43.51 & -66.70 & -0.40 & -0.05 \\
~ & S-Conv \cite{bhalgat2020structured} & -51.23 & -52.18 & -0.63 & -\\
~ & MUSCO \cite{gusak2019automated} & -58.67 & - & -0.47 & -0.30 \\
~ & ADMM-TT \cite{yin2021towards} & -59.51 & - & - & 0.00 \\
~ & CPD-EPC \cite{phan2020stable} & -67.64 & \bf -73.82 & -0.69 & -0.15\\
~ & TRP \cite{xu2019trained} &\bf -68.55 & -3.76 & -2.33 \\
\hline
\multirow{4}*{MobileNetV2} & \bf SVD-NAS & -14.17 & -10.66 & -1.66 & -1.90 \\
\cline{2-6}
~ & Shared \cite{kang2021deeply} & 0.00 & -7.43 & \bf +0.39 & \bf +0.37\\
~ & ALDS \cite{liebenwein2021compressing} & -11.01 & \bf -32.97 & -1.53 & -0.73 \\
~ & S-Conv \cite{bhalgat2020structured} & \bf -19.67 & -25.14 & -0.90 & -\\
\hline
\end{tabular}
\end{center}
\end{table}

\subsection{Ablation Study}
\label{subsection:ablation_study}
In this section, we analyse the individual contribution of each part of our framework.

\subsubsection{Design Space} 
As motioned earlier, although the low-rank approximation problem involves choosing both decomposition scheme and decomposition rank, many existing works \cite{li2019learning,idelbayev2020low} focused on proposing a single decomposition scheme and lowering the rank of the approximation that would minimise the required number of FLOPs with minimum impact on accuracy. 

In our framework, we construct the \textit{LR-space} which expands the space of exploring different decomposition schemes and ranks on a per-layer basis. In Fig.~\ref{fig:random_sample_design_space} Left, the accuracy vs. FLOPs trade-off is plotted for the possible configurations when considering a single layer. As the figure demonstrates, the optimal decomposition scheme and rank depend on the FLOPs allocated to each layer, and the Pareto front is populated with a number of different schemes. These results confirm that previous work which overlooked the choice of decomposition scheme would lead to sub-optimal performance.

\subsubsection{Searching} 
\begin{figure}
    \centering
    \begin{subfigure}{0.24\textwidth}
        \centering
        \includegraphics[width=\textwidth]{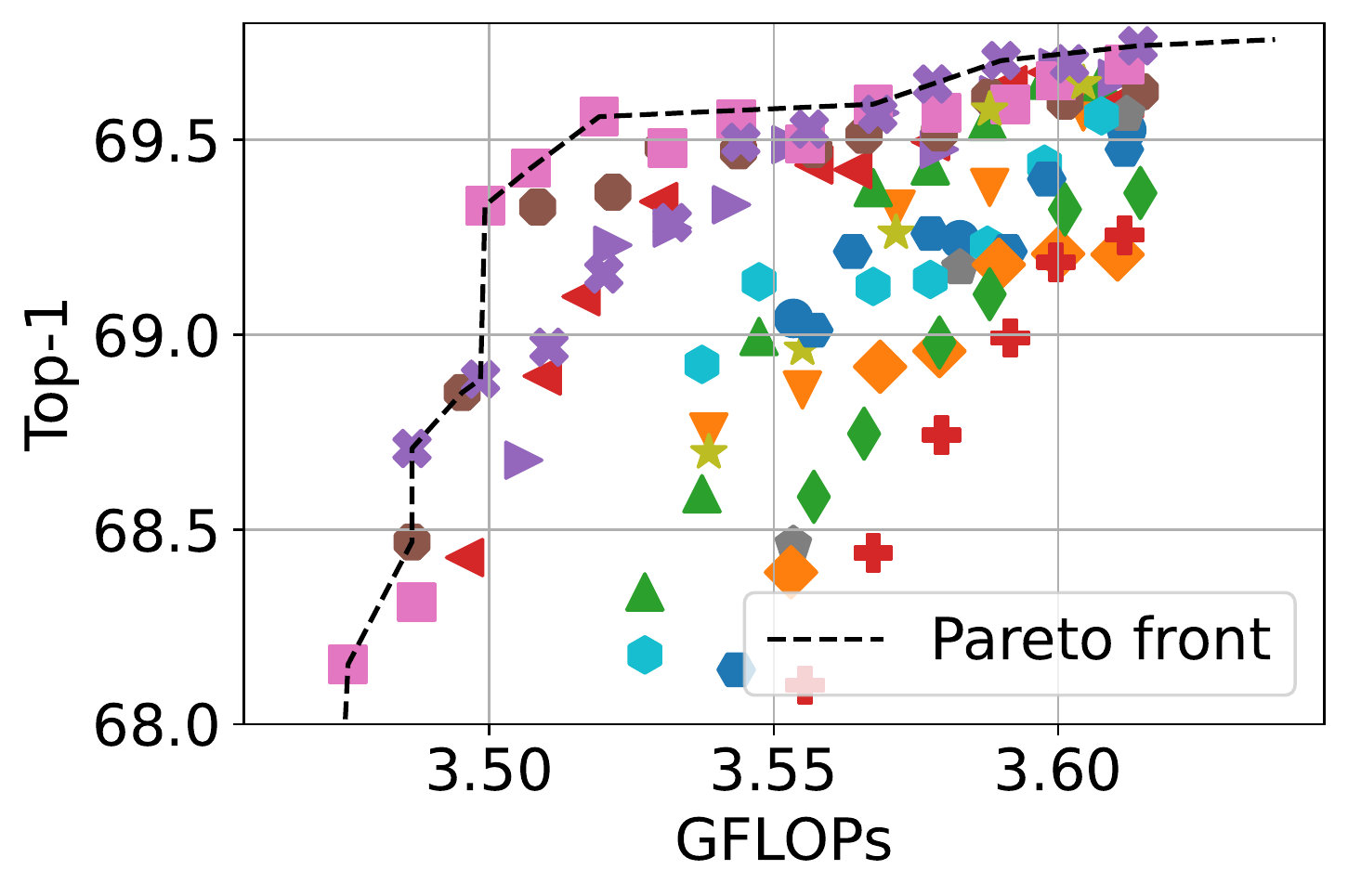}
    \end{subfigure}
    \begin{subfigure}{0.23\textwidth}
        \centering
        \includegraphics[width=\textwidth]{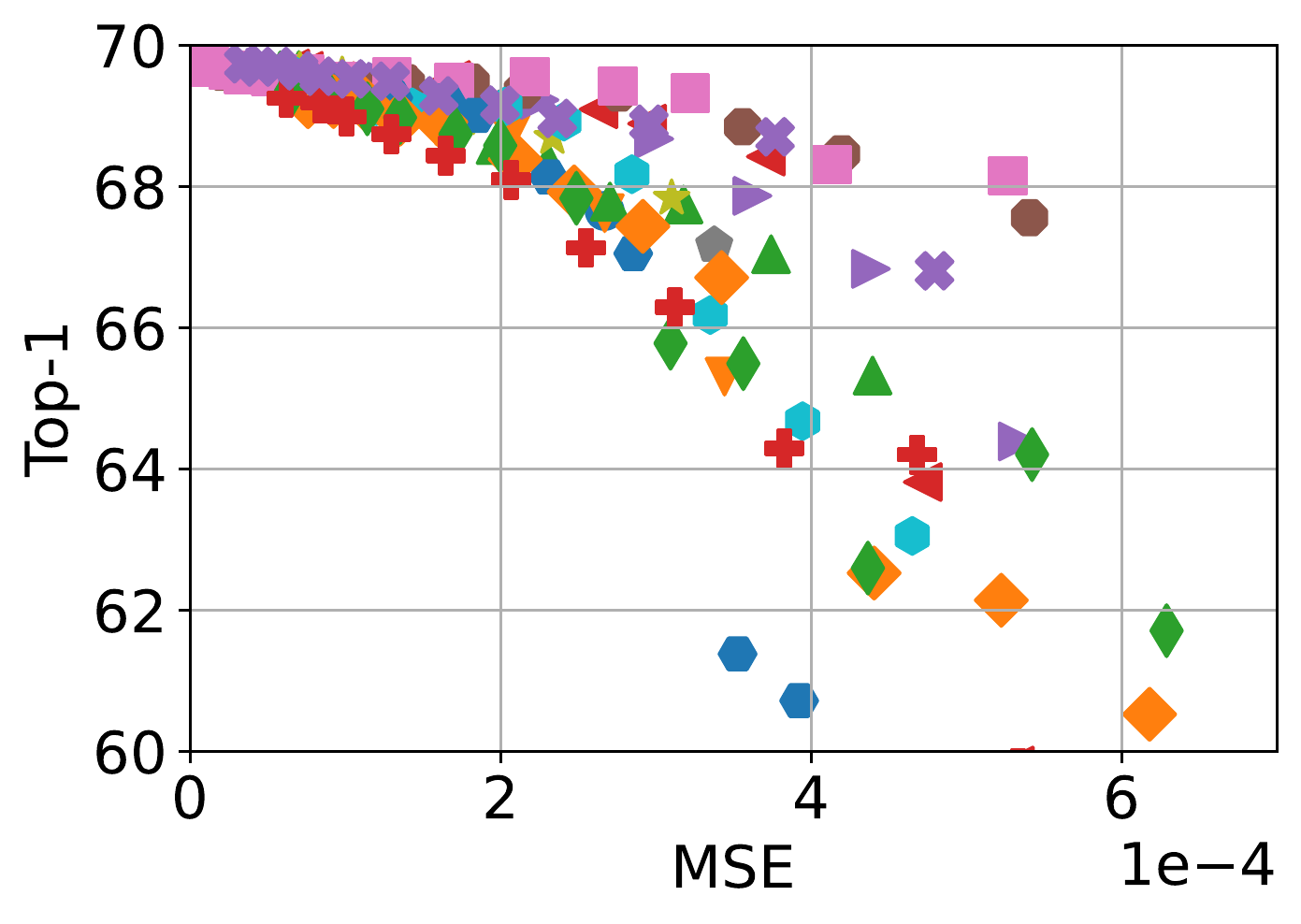}
    \end{subfigure}
    \caption{Explore the \textit{LR-space} of the second convolutional layer in ResNet-18. All other layers are not compressed. Each type of marker corresponds to a specific decomposition scheme. \textbf{Left:} Accuracy versus FLOPs  \textbf{Right:} Accuracy versus MSE}
    \label{fig:random_sample_design_space}
\end{figure}

Many previous works \cite{tai2015convolutional,idelbayev2020low,liebenwein2021compressing} exploit the MSE heuristic to automate the design space exploration of low-rank approximation. Although their methods would result in a faster exploration, that would penalise the quality on estimating the accuracy degradation. Fig.~\ref{fig:random_sample_design_space} Right confirms that MSE of the weight tensor is a poor proxy of the network accuracy degradation. We observed that some configurations of design parameters have similar MSE, but they lead to distinct accuracy results. Therefore, it demonstrates the necessity of using NAS to explore the diverse \textit{LR-space}, which directly optimises accuracy versus FLOPs.

\subsubsection{Synthetic Dataset} 
\begin{figure}[h]
\begin{minipage}{0.28\textwidth}
    \begin{subfigure}{\textwidth}
        \centering
        \includegraphics[width=\textwidth]{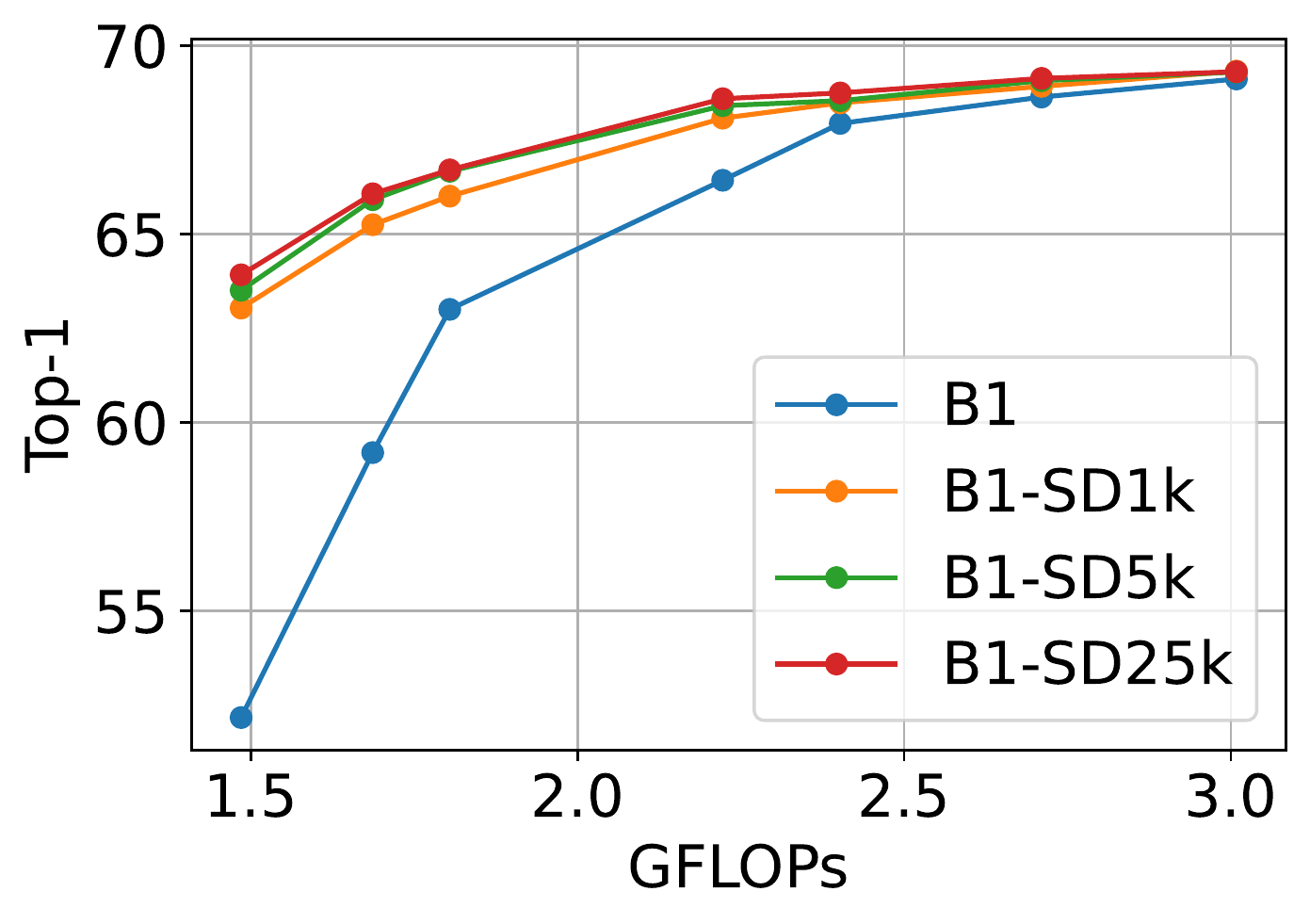}
    \end{subfigure}
\end{minipage} \hfill
\begin{minipage}{0.18\textwidth}
    \begin{subfigure}{\textwidth}
        \centering
        \includegraphics[width=0.5\textwidth]{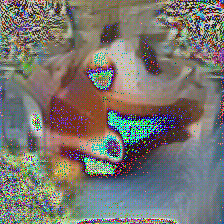}
    \end{subfigure}
    \begin{subfigure}{\textwidth}
        \centering
        \includegraphics[width=0.5\textwidth]{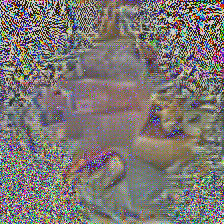}
    \end{subfigure}
\end{minipage}
    \caption{\textbf{Left:} ResNet-18 results of varying the size of synthetic dataset. \textbf{Right Upper and Lower:} Synthetic images of MobileNetV2, taken from our approach and ZeroQ respectively.
    }
    \label{fig:mobilenet_zeroq}
\end{figure}

To investigate the proper quantity of the synthetic images, Fig.~\ref{fig:mobilenet_zeroq} Left demonstrates the top-1 accuracy vs number of FLOPs for 1k, 5k and 25k synthetic images. To distinguish the different experiment configurations that we carried on, they are denoted in the form of $Bx$-$SDy$, where $x$ represents the number of branches in the building block (the residual-style building block is disabled when $x$=1), and $y$ represents the number of synthetic images. The results demonstrate that the accuracy improvement from 5k to 25k images is below 0.5pp. 

\begin{figure*}
    \centering
    \includegraphics[width=0.8\textwidth]{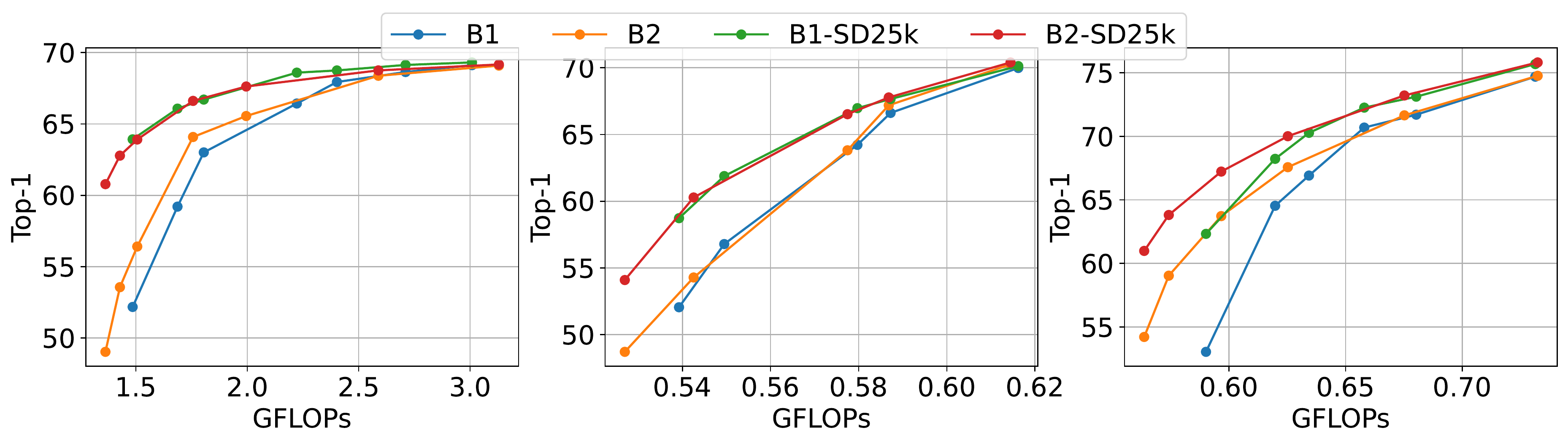}
    \caption{Ablation study of top-1 accuracy-FLOPs trade-off for different configurations of the SVD-NAS approach. \\ \textbf{Left:} ResNet-18. \textbf{Centre:} MobileNetV2. \textbf{Right:} EfficientNet-B0}
    \label{fig:ablation_study}
\end{figure*}

In terms of the quality of the synthetic dataset, although our method is inspired by ZeroQ \cite{cai2020zeroq}, we found their algorithm is not directly applicable to our problem. Compared with ZeroQ, we scale the loss of batch normalisation layers by the number of channels, as \eqref{Eq:bn_loss}. Fig.~\ref{fig:mobilenet_zeroq} Right shows two sample images taken from our method and the original ZeroQ implementation respectively on MobileNet-V2. Without the introduced scaling term, the synthetic image becomes noisy, which we found, is more likely to cause overfitting during the fine-tuning.

\subsubsection{Residual Block} 
An investigation was performed in Fig.~\ref{fig:ablation_study} to assess the impact of multiple branches in the building block of the proposed framework. The system's performance was assessed under considering building blocks with one ($B1$) and two branches ($B2$). In the case of ResNet-18 and EfficientNet-B0, moving from $B1$ to $B2$, improvement in accuracy is obtained, which increases with the compression ratio. 

However, we observed that this gap shrinks when the synthetic dataset is applied, suggesting that the multiple branch flexibility is a more attractive option when we have no training data at all and cannot generate synthetic dataset; in the later case, for example, when the pre-trained model has no batch normalisation layer at all or the batch normalisation layer has already been fused into the convolutional layer.

\subsection{Latency-driven Search}
Till now, the framework has focused on reducing the FLOPs without considering the actual impact on the execution time reduction on a hardware device. We extended the framework by integrating the open-source tool nn-Meter \cite{zhang2021nn} to provide a CNN latency lookup table targeting a Cortex-A76 CPU. The lookup table is then used to replace the FLOPs estimate in \eqref{equ:nas_loss}. Having exposing the latency in the execution of a layer to the framework, we optimised the targeted networks for execution on a Pixel 4 mobile phone. We used a single thread and fixed the batch size to 1. Table \ref{tab:latency} presents the obtained results measured on device, showing that FLOPs can be used as a proxy for performance, especially for ResNet-18 and MobileNetV2. EfficientNet contains SiLU and squeeze-and-excitation operations \cite{tan2019efficientnet}, that currently are not well optimised on CPU and lead to larger discrepancy between latency and FLOPs as a measure of performance.

\begin{table}
\footnotesize
\setlength\tabcolsep{1pt}
\begin{center}
\caption{Latency-driven search results on Pixel 4}
\label{tab:latency}
\begin{tabular}{llllll}
\hline\noalign{\smallskip}
Model & Objective & \makecell{$\Delta$ Top-1\\ (pp)} & \makecell{$\Delta$ FLOPs\\(\%)} & \makecell{$\Delta$ Latency\\(\%)} & \makecell{ Latency\\(ms)}\\
\noalign{\smallskip}
\hline
\noalign{\smallskip}
\multirow{2}*{ResNet-18} & FLOPs & -5.83 & \bf -59.17 & -44.52 &  76.70 \\
~ & Latency & -5.67 & -54.78 & \bf -49.46 & 69.87 \\
\hline
\multirow{2}*{MobileNetV2} & FLOPs & -9.99 & \bf -12.54 & -1.03 & 30.66 \\
~ & Latency & -8.22 & -9.55 & \bf -4.75 & 29.51\\
\hline
\multirow{2}*{EfficientNet-B0} & FLOPs & -9.45 & \bf -22.85 & -1.92 & 67.08 \\
~ & Latency & -10.49 & -21.39 & \bf -6.46 & 63.97 \\
\hline
\end{tabular}
\end{center}
\end{table}

\section{Conclusion}
This paper presents SVD-NAS, a framework that significantly optimises the trade-off between accuracy and FLOPs in data-limited scenarios by fusing the domain of low-rank approximation and NAS. Regarding future work, we will look into further expanding the \textit{LR-space} by including non-SVD based decomposition methods.

\section*{Acknowledgement}
For the purpose of open access, the author(s) has applied a Creative Commons Attribution (CC BY) license to any Accepted Manuscript version arising.

{\small
\bibliographystyle{ieee_fullname}
\bibliography{egbib}

\begin{thebibliography}{10}\itemsep=-1pt

\bibitem{banner2018post}
Ron Banner, Yury Nahshan, Elad Hoffer, and Daniel Soudry.
\newblock Post-training 4-bit quantization of convolution networks for
  rapid-deployment.
\newblock {\em arXiv preprint arXiv:1810.05723}, 2018.

\bibitem{bhalgat2020structured}
Yash Bhalgat, Yizhe Zhang, Jamie~Menjay Lin, and Fatih Porikli.
\newblock Structured convolutions for efficient neural network design.
\newblock {\em Advances in Neural Information Processing Systems},
  33:5553--5564, 2020.

\bibitem{cai2019once}
Han Cai, Chuang Gan, Tianzhe Wang, Zhekai Zhang, and Song Han.
\newblock Once-for-all: Train one network and specialize it for efficient
  deployment.
\newblock {\em arXiv preprint arXiv:1908.09791}, 2019.

\bibitem{cai2018proxylessnas}
Han Cai, Ligeng Zhu, and Song Han.
\newblock Proxylessnas: Direct neural architecture search on target task and
  hardware.
\newblock {\em arXiv preprint arXiv:1812.00332}, 2018.

\bibitem{cai2020zeroq}
Yaohui Cai, Zhewei Yao, Zhen Dong, Amir Gholami, Michael~W Mahoney, and Kurt
  Keutzer.
\newblock Zeroq: A novel zero shot quantization framework.
\newblock In {\em Proceedings of the IEEE/CVF Conference on Computer Vision and
  Pattern Recognition}, pages 13169--13178, 2020.

\bibitem{denton2014exploiting}
Emily Denton, Wojciech Zaremba, Joan Bruna, Yann LeCun, and Rob Fergus.
\newblock Exploiting linear structure within convolutional networks for
  efficient evaluation.
\newblock {\em arXiv preprint arXiv:1404.0736}, 2014.

\bibitem{gusak2019automated}
Julia Gusak, Maksym Kholiavchenko, Evgeny Ponomarev, Larisa Markeeva, Philip
  Blagoveschensky, Andrzej Cichocki, and Ivan Oseledets.
\newblock Automated multi-stage compression of neural networks.
\newblock In {\em Proceedings of the IEEE/CVF International Conference on
  Computer Vision Workshops}, pages 0--0, 2019.

\bibitem{idelbayev2020low}
Yerlan Idelbayev and Miguel~A Carreira-Perpin{\'a}n.
\newblock Low-rank compression of neural nets: Learning the rank of each layer.
\newblock In {\em Proceedings of the IEEE/CVF Conference on Computer Vision and
  Pattern Recognition}, pages 8049--8059, 2020.

\bibitem{jiang2019accuracy}
Weiwen Jiang, Xinyi Zhang, Edwin H-M Sha, Lei Yang, Qingfeng Zhuge, Yiyu Shi,
  and Jingtong Hu.
\newblock Accuracy vs. efficiency: Achieving both through fpga-implementation
  aware neural architecture search.
\newblock In {\em Proceedings of the 56th Annual Design Automation Conference
  2019}, pages 1--6, 2019.

\bibitem{kang2021deeply}
Woochul Kang and Daeyeon Kim.
\newblock Deeply shared filter bases for parameter-efficient convolutional
  neural networks.
\newblock {\em Advances in Neural Information Processing Systems}, 34, 2021.

\bibitem{kim2015compression}
Yong-Deok Kim, Eunhyeok Park, Sungjoo Yoo, Taelim Choi, Lu Yang, and Dongjun
  Shin.
\newblock Compression of deep convolutional neural networks for fast and low
  power mobile applications.
\newblock {\em arXiv preprint arXiv:1511.06530}, 2015.

\bibitem{lazarevich2021post}
Ivan Lazarevich, Alexander Kozlov, and Nikita Malinin.
\newblock Post-training deep neural network pruning via layer-wise calibration.
\newblock {\em arXiv preprint arXiv:2104.15023}, 2021.

\bibitem{lebedev2014speeding}
Vadim Lebedev, Yaroslav Ganin, Maksim Rakhuba, Ivan Oseledets, and Victor
  Lempitsky.
\newblock Speeding-up convolutional neural networks using fine-tuned
  cp-decomposition.
\newblock {\em arXiv preprint arXiv:1412.6553}, 2014.

\bibitem{lee2018snip}
Namhoon Lee, Thalaiyasingam Ajanthan, and Philip~HS Torr.
\newblock Snip: Single-shot network pruning based on connection sensitivity.
\newblock {\em arXiv preprint arXiv:1810.02340}, 2018.

\bibitem{li2018constrained}
Chong Li and CJ Shi.
\newblock Constrained optimization based low-rank approximation of deep neural
  networks.
\newblock In {\em Proceedings of the European Conference on Computer Vision
  (ECCV)}, pages 732--747, 2018.

\bibitem{li2020few}
Tianhong Li, Jianguo Li, Zhuang Liu, and Changshui Zhang.
\newblock Few sample knowledge distillation for efficient network compression.
\newblock In {\em Proceedings of the IEEE/CVF Conference on Computer Vision and
  Pattern Recognition}, pages 14639--14647, 2020.

\bibitem{li2019learning}
Yawei Li, Shuhang Gu, Luc~Van Gool, and Radu Timofte.
\newblock Learning filter basis for convolutional neural network compression.
\newblock In {\em Proceedings of the IEEE/CVF International Conference on
  Computer Vision}, pages 5623--5632, 2019.

\bibitem{liebenwein2021compressing}
Lucas Liebenwein, Alaa Maalouf, Dan Feldman, and Daniela Rus.
\newblock Compressing neural networks: Towards determining the optimal
  layer-wise decomposition.
\newblock {\em Advances in Neural Information Processing Systems}, 34, 2021.

\bibitem{liu2018darts}
Hanxiao Liu, Karen Simonyan, and Yiming Yang.
\newblock Darts: Differentiable architecture search.
\newblock {\em arXiv preprint arXiv:1806.09055}, 2018.

\bibitem{migacz20178}
Szymon Migacz.
\newblock 8-bit inference with tensorrt.
\newblock In {\em GPU technology conference}, volume~2, page~7, 2017.

\bibitem{peng2018extreme}
Bo Peng, Wenming Tan, Zheyang Li, Shun Zhang, Di Xie, and Shiliang Pu.
\newblock Extreme network compression via filter group approximation.
\newblock In {\em Proceedings of the European Conference on Computer Vision
  (ECCV)}, pages 300--316, 2018.

\bibitem{phan2020stable}
Anh-Huy Phan, Konstantin Sobolev, Konstantin Sozykin, Dmitry Ermilov, Julia
  Gusak, Petr Tichavsk{\`y}, Valeriy Glukhov, Ivan Oseledets, and Andrzej
  Cichocki.
\newblock Stable low-rank tensor decomposition for compression of convolutional
  neural network.
\newblock In {\em European Conference on Computer Vision}, pages 522--539.
  Springer, 2020.

\bibitem{sandler2018mobilenetv2}
Mark Sandler, Andrew Howard, Menglong Zhu, Andrey Zhmoginov, and Liang-Chieh
  Chen.
\newblock Mobilenetv2: Inverted residuals and linear bottlenecks.
\newblock In {\em Proceedings of the IEEE conference on computer vision and
  pattern recognition}, pages 4510--4520, 2018.

\bibitem{tai2015convolutional}
Cheng Tai, Tong Xiao, Yi Zhang, Xiaogang Wang, et~al.
\newblock Convolutional neural networks with low-rank regularization.
\newblock {\em arXiv preprint arXiv:1511.06067}, 2015.

\bibitem{tan2019efficientnet}
Mingxing Tan and Quoc Le.
\newblock Efficientnet: Rethinking model scaling for convolutional neural
  networks.
\newblock In {\em International conference on machine learning}, pages
  6105--6114. PMLR, 2019.

\bibitem{tang2020reborn}
Yehui Tang, Shan You, Chang Xu, Jin Han, Chen Qian, Boxin Shi, Chao Xu, and
  Changshui Zhang.
\newblock Reborn filters: Pruning convolutional neural networks with limited
  data.
\newblock In {\em Proceedings of the AAAI Conference on Artificial
  Intelligence}, volume~34, pages 5972--5980, 2020.

\bibitem{wang2022compressing}
Huanyu Wang, Junjie Liu, Xin Ma, Yang Yong, Zhenhua Chai, and Jianxin Wu.
\newblock Compressing models with few samples: Mimicking then replacing.
\newblock {\em arXiv preprint arXiv:2201.02620}, 2022.

\bibitem{wang2017factorized}
Min Wang, Baoyuan Liu, and Hassan Foroosh.
\newblock Factorized convolutional neural networks.
\newblock In {\em Proceedings of the IEEE International Conference on Computer
  Vision Workshops}, pages 545--553, 2017.

\bibitem{wu2019fbnet}
Bichen Wu, Xiaoliang Dai, Peizhao Zhang, Yanghan Wang, Fei Sun, Yiming Wu,
  Yuandong Tian, Peter Vajda, Yangqing Jia, and Kurt Keutzer.
\newblock Fbnet: Hardware-aware efficient convnet design via differentiable
  neural architecture search.
\newblock In {\em Proceedings of the IEEE/CVF Conference on Computer Vision and
  Pattern Recognition}, pages 10734--10742, 2019.

\bibitem{xu2019trained}
Yuhui Xu, Yuxi Li, Shuai Zhang, Wei Wen, Botao Wang, Wenrui Dai, Yingyong Qi,
  Yiran Chen, Weiyao Lin, and Hongkai Xiong.
\newblock Trained rank pruning for efficient deep neural networks.
\newblock In {\em 2019 Fifth Workshop on Energy Efficient Machine Learning and
  Cognitive Computing-NeurIPS Edition (EMC2-NIPS)}, pages 14--17. IEEE, 2019.

\bibitem{yin2021towards}
Miao Yin, Yang Sui, Siyu Liao, and Bo Yuan.
\newblock Towards efficient tensor decomposition-based dnn model compression
  with optimization framework.
\newblock In {\em Proceedings of the IEEE/CVF Conference on Computer Vision and
  Pattern Recognition}, pages 10674--10683, 2021.

\bibitem{yu2017compressing}
Xiyu Yu, Tongliang Liu, Xinchao Wang, and Dacheng Tao.
\newblock On compressing deep models by low rank and sparse decomposition.
\newblock In {\em Proceedings of the IEEE Conference on Computer Vision and
  Pattern Recognition}, pages 7370--7379, 2017.

\bibitem{zhang2021nn}
Li~Lyna Zhang, Shihao Han, Jianyu Wei, Ningxin Zheng, Ting Cao, Yuqing Yang,
  and Yunxin Liu.
\newblock Nn-meter: Towards accurate latency prediction of deep-learning model
  inference on diverse edge devices.
\newblock In {\em Proceedings of the 19th Annual International Conference on
  Mobile Systems, Applications, and Services}, pages 81--93, 2021.

\bibitem{zhang2015accelerating}
Xiangyu Zhang, Jianhua Zou, Kaiming He, and Jian Sun.
\newblock Accelerating very deep convolutional networks for classification and
  detection.
\newblock {\em IEEE transactions on pattern analysis and machine intelligence},
  38(10):1943--1955, 2015.

\bibitem{zhang2015efficient}
Xiangyu Zhang, Jianhua Zou, Xiang Ming, Kaiming He, and Jian Sun.
\newblock Efficient and accurate approximations of nonlinear convolutional
  networks.
\newblock In {\em Proceedings of the IEEE Conference on Computer Vision and
  pattern Recognition}, pages 1984--1992, 2015.

\end{thebibliography}
}
\clearpage 

\setcounter{page}{1}
\setcounter{section}{0}
\setcounter{table}{0}
\setcounter{figure}{0}
\setcounter{equation}{0}
\setcounter{algorithm}{0}

\renewcommand{\thepage}{S\arabic{page}} 
\renewcommand{\thesection}{S\arabic{section}}  
\renewcommand{\thetable}{S\arabic{table}}  
\renewcommand{\thefigure}{S\arabic{figure}}
\renewcommand{\theequation}{S.\arabic{equation}}
\renewcommand{\thealgorithm}{S.\arabic{algorithm}}


\section{Implementation Details}
\subsection{Identify the Optimal Design Point}
With regards to which layers to compress, for ResNet-18, we target every $3\times3$ convolutional layer except the first layer. For MobileNetV2 and EfficientNet-B0, we compress all the point-wise convolutions. 

To reduce the searching cost, the design space is pruned by FLOPs for all three models, where the step size of the grid search is 5\%. When pruning by accuracy, 500 images are randomly selected from the validation set for the proxy task, and these images are fixed throughout the experiment. This technique is only used on MobileNetV2 and EfficientNet-B0, and the accuracy degradation tolerance of the proxy task $\tau_{proxy}$ is set as 5pp of top-1 accuracy. 

During NAS, the sampling parameters $\boldsymbol{\theta_{i}}$ are learned using Adam with $lr$=0.01, $betas$=(0.9,0.999), $weight\;decay$=0.0005. The learning process takes 100 epochs and 50 epochs respectively for searching the first and the second branch. The Gumbel-Softmax temperature is initialised as 5, and decays by 0.965 for every epoch.

\subsection{Iterative Searching Method}
The proposal of the iterative searching method aims to reduce the searching cost by focusing on one branch of the building block at each time. Since this method identifies the optimal configuration of each branch in a greedy way, it is at the risk of not finding the global optimal solution. Empirically, we found that relaxing the configuration identified for the first branch can help us to reach better design points.

As Algorithm~\ref{alg:Iterative_Searching_with_Relaxed_Configuration} shows, for the first branch, we choose to relax its configuration from $r_{i}^{0}$ to $r_{i}^{0,*}$ by removing a proportion of ranks $\Delta r_{i}$. The intuition behind is that the relaxation will give the second iteration of NAS more space to explore, probably reducing the chance of being stuck at local optimum. In our experiment, we choose the value of $\Delta r_{i}$ so that the corresponding FLOPs difference between using $r_{i}^{0}$ and $r_{i}^{0,*}$ is equal to 20\% FLOPs of the \textit{original layer}.

\begin{algorithm}[h]
\caption{Iterative Searching (Relaxed Configuration)}
\label{alg:Iterative_Searching_with_Relaxed_Configuration}
\small
\begin{algorithmic}[1]
\STATE $\boldsymbol{E_{i,0}} = \boldsymbol{W_{i}}$
\FOR{$b \in \{0, 1\}$}
\STATE identify optimal $F_{b}(\boldsymbol{\hat{W}_{i}^{1,b}}, \boldsymbol{\hat{W}_{i}^{0,b}})$ to approximate $\boldsymbol{E_{i,b}}$, with the rank of $r_{i}^{b}$
\IF{b==0}
    \STATE \textcolor{blue}{relax the rank $r_{i}^{b,*}$ = $r_{i}^{b}$ - $\Delta r_{i}$, and update the low-rank weight tensors accordingly}
\ENDIF
\STATE $\boldsymbol{E_{i,b+1}} = \boldsymbol{E_{i,b}}-F_{b}(\boldsymbol{\hat{W}_{i}^{1,b}}, \boldsymbol{\hat{W}_{i}^{0,b}})$
\ENDFOR
\end{algorithmic}
\end{algorithm}

\subsection{Generate the Synthetic Dataset}
Each random initialised image is optimised with Adam for 500 iterations. The learning rate is initialised at 0.5, 0.25 and 0.5 for ResNet-18, MobileNetV2 and EfficientNet-B0 respectively, and it is scheduled to decay by 0.1 as long as the loss stopped improving for 100 iterations. $\alpha$ in $l_{bn}$ is set to 1 for ResNet-18 and MobileNetV2, and 100 for EfficientNet-B0. The scaling factor $\frac{1}{f_{i}}$ is introduced on MobileNetV2 and EfficientNet-B0 only, but not ResNet-18. The batch size is set to 32 for all three models. 

Compared with ZeroQ, Our objective function $l_{bn}$ differs in that we averaged the loss for each batch normalisation layer by scaling it with the number of channels $f_{i}$. We found taking this average is important for those compact models which use MBBlock, such as MobileNetV2 and EfficientNet to produce high-quality synthetic images. The intuition behind this change is that the number of channels in these compact models varies a lot between layers because of the inverted residual structure. Therefore, averaging by $f_{i}$ helps balancing the contribution of each batch normalisation layer towards the total loss.

In terms of the storage of the synthetic dataset, the floating point format leads to better knowledge distillation results the quantised RGB format. As such, a synthetic dataset containing 25k images requires about 14 GB disk space.

\subsection{Fine-tune the Low-rank Model}
For all the experiment set-ups: post-training, few-sample training and full training, our framework uses SGD for fine-tuning, with momentum and weight decay set to 0.9 and 0.0001 respectively. The learning rate is initialised to 0.001 and set to decay by 0.1, as long as validation accuracy has
not been improved for the last 10 epochs. The fine-tuning stops
once the learning rate is below 0.0001. 

\subsection{Measure Performance}
\subsubsection{FLOPs and Parameters Measurement} The number of FLOPs and parameters in a given model is calculated using the open-source library thop\footnote{https://github.com/Lyken17/pytorch-OpCounter}, and the number of FLOPs is defined as the twice of Multiply-Accumulate (MAC). 

\subsubsection{Latency Measurement} The reported CPU latency is measured on a Pixel 4 phone using the Tensorflow Lite 2.1 native benchmark binary \footnote{https://www.tensorflow.org/lite/performance/measurement}. The CNN model runs on a single thread using one of the Cortex-A76 core, and the reported result is the average of 50 runs.

\section{Reproduce Previous Work}
When comparing our SVD-NAS with existing works, for few-sample training and full training set-ups, we presented the data reported in those papers. However, for the post-training set-up, those relevant works did not disclose the precise data. Therfore, we reproduce the following works to generate the numbers in the table of post-training comparison. The code for reproducing these works has also been released.

\subsection{ALDS} The authors of ALDS demonstrated the post-training results, referred to as ``Compress-only" in their paper, in figures without providing the actual performance numbers. Therefore, we use their official code \footnote{https://github.com/lucaslie/torchprune} to rerun the post-training tests. For fair comparison, we altered their way of FLOPs and parameters measurement to align with ours, as their official code only counts the FLOPs in the convolutional and fully-connected layers rather than the entire model, which would overestimate the compression ratio.

\subsection{LR-S2} The authors of LR-S2 proposed a single decomposition scheme and a rank selection method, which is:
\begin{equation}
    \min_{r_{k}}(\sum_{k=1}^{K}( \lambda C_{k}(r_{k})+\frac{\mu}{2}\sum_{i=r_{k}+1}^{R_{k}} s_{k,i}^{2})).
    \label{eq:learn_s2}
\end{equation}
$K$ denotes the number of layers in the model, and $r_{k}$ denotes the decomposition rank in the $k^{th}$ layer. In addition, $C_{k}(r_{k})$ denotes the per-layer computation cost in FLOPs and $s_{k,i}$ denotes the $i^{th}$ largest singular value in the $k^{th}$ layer. $\lambda$ and $\mu$ are two hyperparameters.

However, since the authors of LR-S2 did not consider the data-limited setting, they chose to use \eqref{eq:learn_s2} as a regularisation term and solve it by gradient descent. Instead, we solved the above optimisation problem without the need of data by identifying the minimal rank $r_{k}$ which makes the following inequality hold true for each layer.

\begin{equation}
    \lambda C_{k}(r_{k}+1) - \lambda C_{k}(r_{k}) - \frac{\mu}{2}s_{k,r_{k}+1}^{2} \geq 0
\end{equation}

\subsection{F-Group} F-Group is a work which manually picked the decomposition scheme as well as the decomposition rank. We reproduced the exact same decomposition configuration shown in the paper and collected the model accuracy without any fine-tuning. We observed this hard-crafted method heavily degrades the model accuracy to almost zero top-1 accuracy.

\begin{figure*}
    \centering
    \includegraphics[width=0.8\textwidth]{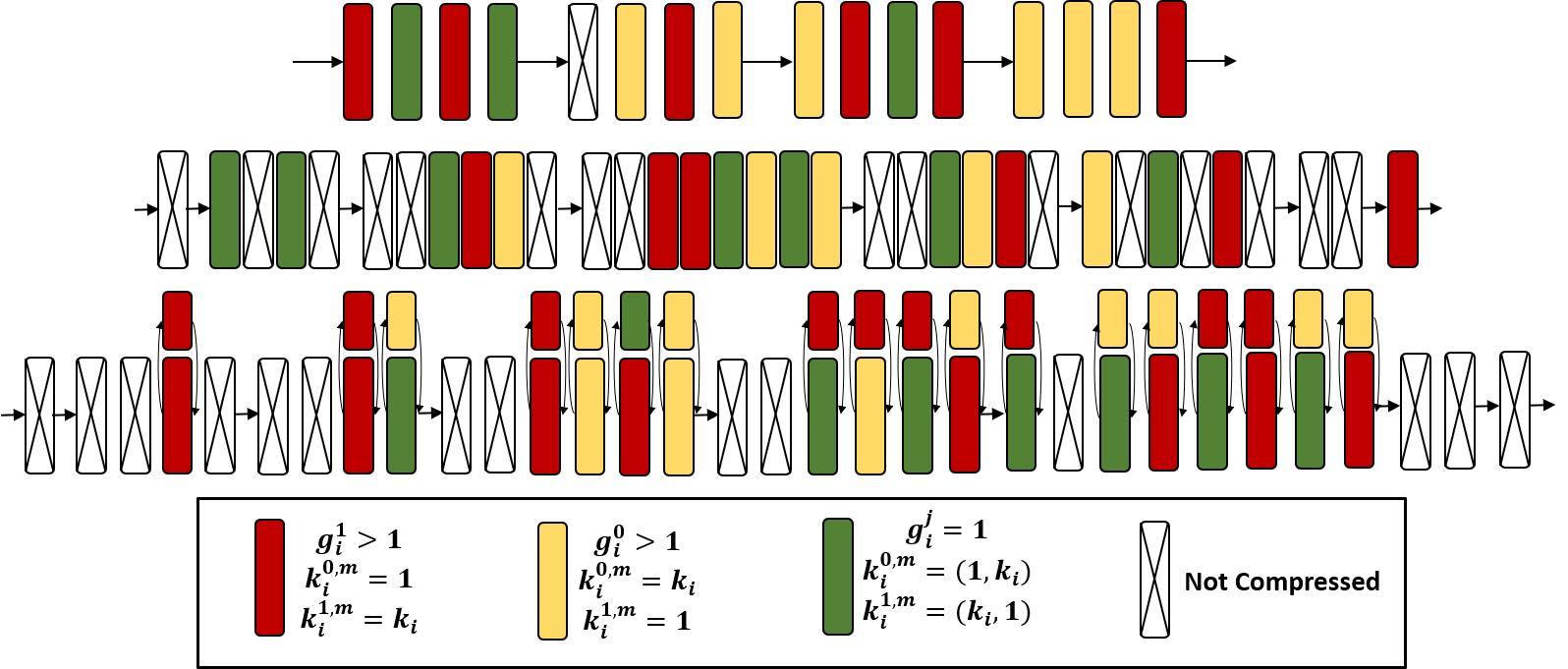}
    \caption{Schematics of the obtained low-rank models. The coloured rectangles represent SVD building blocks and they are clustered into three types (red, yellow, green) based on the value of their design parameters.
    \textbf{Top:} ResNet-18, single-branch design, obtained by the setting that $\beta$=48, $\gamma$=[0.3,0.7].
    \textbf{Middle:} MobileNetV2, single-branch design, the setting is $\beta$=88, $\gamma$=[0.6,0.95], $\tau_{proxy}$=5.
    \textbf{Bottom:} EfficientNet-B0, two-branch design, the setting is $\beta$=48, $\gamma$=[0.5,0.95], $\tau_{proxy}$=5.}
    \label{fig:model_visualise}
\end{figure*}

\section{Additional Results}
\subsection{Searching Time}
In our framework, the procedure of NAS is completed on a single RTX 2080 Ti or a GTX 1080 Ti GPU. Table~\ref{tab:searching_time} summarises the averaged searching time which mainly includes deriving the weights of SVD building blocks, pruning the design space, and the training loop of the sampling parameters.

\setlength{\tabcolsep}{4pt}
\begin{table}[h]
\footnotesize
\begin{center}
\caption{Averaged execution time of NAS.}
\label{tab:searching_time}
\begin{tabular}{lllll}
\hline\noalign{\smallskip}
\multirow{2}*{Model} & \multicolumn{2}{c}{\underline{first iteration}} & \multicolumn{2}{c}{\underline{second iteration}}\\
~ & GPU & hours & GPU & hours \\
\noalign{\smallskip}
\hline
\noalign{\smallskip}
ResNet-18 & 2080 & 8.66 & 1080 & 4.66 \\
\hline
MobileNetV2 & 1080 & 10.03 & 1080 & 3.07 \\
\hline
EfficientNet-B0 & 1080 & 14.67 & 1080 & 3.07 \\
\hline
\end{tabular}
\end{center}
\end{table}

\subsection{Overfitting in the Search}
Due to the data-limited problem setting that the access to training data is difficult, the sampling parameters $\boldsymbol{\theta_{i}}$ are learned on the validation set. To understand the effect of overfitting, Table~\ref{tab:additional_overfitting} compares the performance of the identified design point when searching with the whole validation set and 1\% of it. The results show that, under the similar compression ratio, the accuracy difference is no more than 1.07pp.

\setlength{\tabcolsep}{2pt}
\begin{table}[h]
\footnotesize
\begin{center}
\caption{Train the sampling parameters $\boldsymbol{\theta_{i}}$ using the whole (50k samples) and 1\% (500 samples) of ImageNet validation set. Experiment set-up is B1-SD25k.}
\label{tab:additional_overfitting}
\begin{tabular}{llllll}
\hline\noalign{\smallskip}
Model & Val Size & \makecell{$\Delta$ FLOPs \\ (\%)} & \makecell{$\Delta$ Params\\ (\%)}  & \makecell{$\Delta$ Top-1\\ (pp)} & \makecell{$\Delta$ Top-5\\ (pp)}\\
\noalign{\smallskip}
\hline
\noalign{\smallskip}
\multirow{2}*{ResNet-18} & 50k & -59.17 & -66.77 & -5.83 & -3.39 \\
~ & 500 & -58.96 & -67.57 & -6.90 & -3.95 \\
\hline
\multirow{2}*{MobileNetV2} & 50k & -12.54 & -9.00 & -9.99 & -6.11 \\
~ & 500 & -12.37 & -8.73 & -10.79 & -6.53  \\
\hline
\multirow{2}*{EfficientNet-B0} & 50k & -26.53 & -17.69 & -15.35 & -8.99 \\
~ & 500 & -26.28 & -17.53 & -13.53 & -7.64  \\
\hline
\end{tabular}
\end{center}
\end{table}

\subsection{Visualise Low-rank Models}
In order to visualise the optimal design parameters identified by the gradient-descent NAS, we sketched several examples of the obtained low-rank models in Fig.~\ref{fig:model_visualise}. As our framework targets the compression of every $3\times3$ convolutions except the first layer in ResNet-18, and every point-wise convolutions in MobileNetV2 and EfficientNet-B0, only these targeted layers are demonstrated while others are hidden from the schematics. These visualised results demonstrate the advantage of our framework over the previous work that we have a larger design space and we can also efficiently explore that space.

\subsection{Complete Post-training Results}
Table~\ref{tab:complete_post_training} provides the complete post-training results of SVD-NAS, including metrics of interest and the corresponding hyperparameters to obtain these designs. 

\setlength{\tabcolsep}{4pt}
\begin{table*}
\footnotesize
\begin{center}
\caption{Post-training results of SVD-NAS. $\Delta$ F/P denote the FLOPs and parameters reduction in percentage (\%), while $\Delta$ Top1/5 denote the accuracy degradation in percentage point (pp). $\beta$ is the hyperparameter which balances the cross-entropy term and the computation cost term in the loss function of gradient descent NAS. $\gamma$ and $\tau_{proxy}$ control the level of design space pruning. When pruning by FLOPs, $\gamma$ denotes the values of the grid search, with a default step size at 0.05; for example, $\gamma$=[0.3,0.7] means ``0.3, 0.35, 0.4 ... 0.7". When pruning by accuracy, $\tau_{proxy}$ denotes the tolerance of top-1 accuracy degradation in the units of pp.}
\label{tab:complete_post_training}
\begin{tabular}{l|ll|ll|ll}
\hline\noalign{\smallskip}
\multirow{2}*{ResNet-18} & \multicolumn{2}{c|}{$\beta$=48, $\gamma$=[0.3,0.7]} & \multicolumn{2}{c|}{$\beta$=48, $\gamma$=[0.4,0.8]} & \multicolumn{2}{c}{$\beta$=32, $\gamma$=[0.4,0.8]}  \\
~ & $\Delta$ F/P & $\Delta$ Top1/5 & $\Delta$ F/P & $\Delta$ Top1/5 & $\Delta$ F/P  & $\Delta$ Top1/5 \\
\noalign{\smallskip}
\hline
\noalign{\smallskip}
 B1 & \multirow{3}*{-59.17/-66.77} & -17.58/-12.79 & \multirow{3}*{-53.64/-59.79} & -10.55/-7.22 & \multirow{3}*{-50.39/-58.79} & -6.75/-4.41 \\
 B1-SD1k & ~  & -6.72/-4.14 & ~ & -4.50/-2.48 & ~ & -3.75/-2.10 \\
 B1-SD25k & ~ & -5.83/-3.39 & ~ & -3.68/-2.03 & ~ & -3.05/-1.62 \\
\hline
 B2 & \multirow{2}*{-62.53/-72.61} & -20.73/-18.88 & \multirow{2}*{-60.74/-69.21} & -16.20/-11.12 & \multirow{2}*{-58.60/-68.05} & -13.35/-9.14\\
 B2-SD25k & ~ & -8.98/-5.28 & ~ & -6.98/-4.07 & ~ & -5.85/-3.34 \\
 \hline\noalign{\smallskip}
\multirow{2}*{ResNet-18} & \multicolumn{2}{c|}{$\beta$=16, $\gamma$=[0.5,0.9]} & \multicolumn{2}{c|}{$\beta$=4, $\gamma$=[0.5,0.9]} & \multicolumn{2}{c}{$\beta$=2, $\gamma$=[0.5,0.9]} \\
~ & $\Delta$ F/P & $\Delta$ Top1/5 & $\Delta$ F/P  & $\Delta$ Top1/5 & $\Delta$ F/P  & $\Delta$ Top1/5 \\
\noalign{\smallskip}
\hline
\noalign{\smallskip}
 B1 & \multirow{3}*{-38.91/-43.71} & -3.32/-2.5 & \multirow{3}*{-25.50/-36.70} & -1.12/-0.84 & \multirow{3}*{-17.30/-27.19} & -0.64/-0.24\\
 B1-SD1k & ~ & -1.68/-0.85 & ~ & -0.83/-0.36 & ~ & -0.43/-0.21 \\
B1-SD25k & ~ & -1.16/-0.61 & ~ & -0.62/-0.27 & ~ & -0.44/-0.09\\
 \hline
 B2 & \multirow{2}*{-45.16/-53.64} & -4.20/-2.52 & \multirow{2}*{-28.86/-42.19} & -1.38/-0.75 & \multirow{2}*{-13.98/-26.94} & -0.67/-0.30 \\
 B2-SD25k & ~ & -2.13/-0.95 & ~ & -1.00/-0.38 & ~ & -0.59/-0.18 \\
\hline\noalign{\smallskip}
\hline\noalign{\smallskip}
\multirow{2}*{MobileNetV2} & \multicolumn{2}{c|}{$\beta$=88, $\gamma$=[0.6,0.95], $\tau_{proxy}$=5} & \multicolumn{2}{c|}{$\beta$=80, $\gamma$=[0.6,0.95], $\tau_{proxy}$=5} & \multicolumn{2}{c}{$\beta$=64, $\gamma$=[0.6,0.95], $\tau_{proxy}$=5}  \\
~ & $\Delta$ F/P & $\Delta$ Top1/5 & $\Delta$ F/P  & $\Delta$ Top1/5 & $\Delta$ F/P  & $\Delta$ Top1/5  \\
\noalign{\smallskip}
\hline
\noalign{\smallskip}
B1 & \multirow{3}*{-14.17/-10.66} & -19.82/-13.15  & \multirow{3}*{-12.54/-9.00} & -15.09/-7.79 & \multirow{3}*{-7.73/-6.70} & -7.66/-4.46\\
B1-SD1k & ~ & -14.41/-9.00 & ~ & -11.63/-7.17 & ~ & -5.51/-3.10 \\
B1-SD25k & ~ & -13.15/-8.34 & ~ & -9.99/-6.11 & ~ & -4.91/-2.88 \\
\hline
B2 & \multirow{2}*{-16.13/-13.24} & -23.16/-15.75 & \multirow{2}*{-13.64/-9.70} & -17.59/-11.39 & \multirow{2}*{-8.09/-6.60} & -8.06/-4.70\\
 B2-SD25k & ~ & -17.78/-11.35 & ~ & -11.60/-6.92 & ~ & -5.36/-2.97 \\
\hline\noalign{\smallskip}
\multirow{2}*{MobileNetV2} & \multicolumn{2}{c|}{$\beta$=48, $\gamma$=[0.6,0.95], $\tau_{proxy}$=5} & \multicolumn{2}{c|}{$\beta$=32, $\gamma$=[0.6,0.95], $\tau_{proxy}$=5} & & \\
 ~ & $\Delta$ F/P & $\Delta$ Top1/5 & $\Delta$ F/P  & $\Delta$ Top1/5 & &\\
\noalign{\smallskip}
\hline
\noalign{\smallskip}
B1 & \multirow{3}*{-6.53/-5.82} & -5.26/-3.08 & \multirow{3}*{-1.92/-2.74} & -1.90/-0.86\\
B1-SD1k &  ~ & -4.41/-2.52 & ~ & -1.89/-0.84 \\
B1-SD25k & ~ & -4.22/-2.35 & ~ & -1.76/-0.73 \\
\hline
B2 & \multirow{2}*{-6.60/-5.71} & -4.69/-2.72 & \multirow{2}*{-2.20/-2.34} & -1.62/-0.84\\
 B2-SD25k & ~ & -4.10/-2.21 & ~ & -1.48/-0.78\\
\hline\noalign{\smallskip}
\hline\noalign{\smallskip}
\multirow{2}*{EfficientNet-B0} & \multicolumn{2}{c|}{$\beta$=88, $\gamma$=[0.5,0.95], $\tau_{proxy}$=5} & \multicolumn{2}{c|}{$\beta$=80, $\gamma$=[0.5,0.95], $\tau_{proxy}$=5} & \multicolumn{2}{c}{$\beta$=64, $\gamma$=[0.5,0.95], $\tau_{proxy}$=5}  \\
~ & $\Delta$ F/P & $\Delta$ Top1/5 & $\Delta$ F/P  & $\Delta$ Top1/5 & $\Delta$ F/P  & $\Delta$ Top1/5  \\
\noalign{\smallskip}
\hline
\noalign{\smallskip}
B1 & \multirow{3}*{-26.53/-17.69} & -24.65/-15.92 & \multirow{3}*{-22.85/-16.06} & -13.15/-7.45 & \multirow{3}*{-21.04/-15.60} & -10.77/-5.95\\
B1-SD1k & ~ & -16.36/-9.62 & ~ & -9.68/-5.18 & ~ & -7.48/-3.90\\
B1-SD25k & ~ & -15.35/-8.99 & ~ & -9.45/-5.08 & ~ & -7.40/-3.78\\
\hline
B2 & \multirow{2}*{-29.83/-20.60} & -23.47/-15.02 & \multirow{2}*{-28.52/-20.38} & -18.65/-11.20 & \multirow{2}*{-25.72/-18.86} & -13.96/-8.10 \\
 B2-SD25k & ~ & -16.70/-10.01 & ~ & -13.87/-7.87 & ~ & -10.45/-5.69\\
\hline\noalign{\smallskip}
\multirow{2}*{EfficientNet-B0} & \multicolumn{2}{c|}{$\beta$=48, $\gamma$=[0.5,0.95], $\tau_{proxy}$=5} & \multicolumn{2}{c|}{$\beta$=32, $\gamma$=[0.5,0.95], $\tau_{proxy}$=5} & \multicolumn{2}{c}{$\beta$=16, $\gamma$=[0.5,0.95], $\tau_{proxy}$=5}  \\
 ~ & $\Delta$ F/P & $\Delta$ Top1/5 & $\Delta$ F/P  & $\Delta$ Top1/5 & $\Delta$ F/P  & $\Delta$ Top1/5 \\
\noalign{\smallskip}
\hline
\noalign{\smallskip}
B1 & \multirow{3}*{-18.10/-13.23} & -6.99/-3.79  & \multirow{3}*{-15.32/-12.08} & -5.98/-3.05  & \multirow{3}*{-8.97/-6.11} & -2.98/-1.52\\
B1-SD1k & ~ & -5.60/-2.76 & ~ & -4.65/-2.36 & ~ & -2.09/-0.98 \\
B1-SD25k & ~ & -5.42/-2.79 & ~ & -4.56/-2.37 & ~ & -1.99/-0.87 \\
\hline
B2 & \multirow{2}*{-22.17/-16.41} & -10.11/-5.49 & \multirow{2}*{-15.95/-12.85} & -6.03/-3.16 & \multirow{2}*{-8.85/-6.19} & -2.91/-1.54 \\
 B2-SD25k & ~ & -7.67/-4.06 & ~ & -4.46/-2.32 & ~ & -1.86/-0.91 \\
\hline
\end{tabular}
\end{center}
\end{table*}

\end{document}